\documentclass{sig-alternate}
\usepackage{caption}
\usepackage{verbatim}
\usepackage{subcaption}
\usepackage{comment}
\newtheorem{definition}{Definition}
\newtheorem{observation}{Observation}

\newtheorem{problem}{Problem}
\newtheorem{theorem}{Theorem}

\newtheorem{result}{Result}
\usepackage{epstopdf}
\begin{document}

\conferenceinfo{WOODSTOCK}{'97 El Paso, Texas USA}

\title{ Sleep Analytics and Online Selective Anomaly Detection
}

%
%
%
%
%

\numberofauthors{3} 
%
\author{
%
%
\alignauthor
Tahereh Babaie\\
       \affaddr{School of IT}\\
       \affaddr{University of Sydney}\\
       \affaddr{Sydney, NSW, Australia}\\
       \email{\fontsize{10}{12} \selectfont tahereh.babaie@sydney.edu.au}
\alignauthor
Sanjay Chawla\\
       \affaddr{School of IT}\\
       \affaddr{University of Sydney}\\
        \affaddr{Sydney, NSW, Australia}\\
       \email{\fontsize{10}{12} \selectfont sanjay.chawla@sydney.edu.au}
\alignauthor
Romesh Abeysuriya\\
       \affaddr{School of Physics}\\
       \affaddr{University of Sydney}\\
        \affaddr{Sydney, NSW, Australia}\\
       \email{\fontsize{10}{12} \selectfont r.abeysuriya@physics.usyd.edu.au}
}

\maketitle
\begin{abstract}
We introduce a new problem, the Online Selective Anomaly Detection (OSAD),
to model a specific scenario emerging from research in sleep science.
Scientists have segmented sleep into several stages and stage two
is characterized by two patterns (or anomalies) in the EEG time series
recorded on sleep subjects. These two patterns are sleep spindle (SS)
and K-complex. The OSAD problem was introduced to design a residual
system, where all anomalies (known and unknown) are detected but
the system only triggers an alarm when non-SS anomalies appear.
The solution of the OSAD problem required us to combine
techniques from both machine learning and control theory. Experiments
on data from real subjects attest to the effectiveness of our approach.

\end{abstract}


\terms{Anomaly/novelty detection, Mining rich data types}


\section{Introduction}
Research in human sleep condition has emerged as a rapidly growing
area within medicine, biology and physics. A defining aspect of
sleep research is the large amount of data that is generated
in a typical sleep experiment.

A sleep experiment consists of a  human subject, in a state of sleep,
whose neural activity is being recorded with Electroencephalography (EEG) \cite{Niedermeyer2005, Buzsaki2006}. A typical
full night EEG time-series, recorded between 4-64 locations
on the scalp, at 200 Hz, for eight hours, will generate approximately
300MB of data. A typical clinical study will have between ten and
fifty subjects. Surprisingly vast majority of sleep clinics still
use a manual process to analyze the recorded EEG time-series. Hence
there is considerable interest in automating the analysis of EEG generated
from sleep experiments.
\begin{figure} [ht!]
        \hspace{-30pt}
         \includegraphics[width=0.6\textwidth]{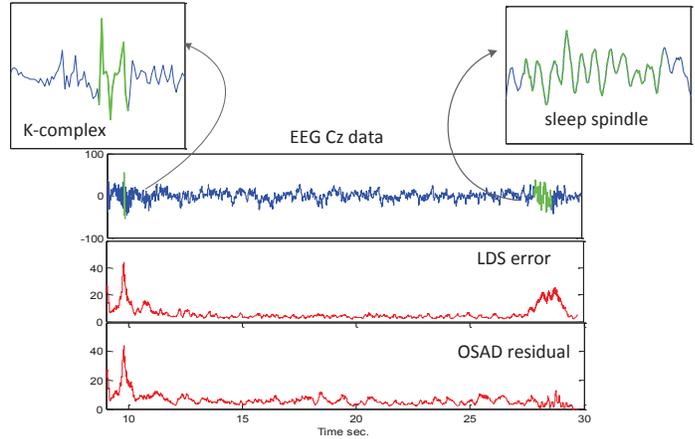}
         \vspace{-20pt}
         \caption{Sleep spindles (SS) along with K-Complexes (KC) are defining characteristics of stage 2 sleep. Both SS and KC will show up as residuals in
an LDS system. The OSAD problem will lead to a {\em new} residual time-series
where SS will be automatically supressed but KC will remain unaffected.
Due to relatively high frequency of SS, there are certain situations
where sleep scientists only want to be alerted when a non-SS anomaly occurs}
\label{fig:example}
\end{figure}
Scientists have segmented sleep into several stages based on the
responsiveness of the subject and other physiological features. Of
particular important is what is termed as stage 2 (moderately deep
sleep). This stage is characterized by two  phenomenon that occur
in the EEG time series.  These are {\em sleep spindles}, which are transient bursts of neural activity with a characteristic frequency of 12--14\,Hz, and
{\em K-Complexes}, which are short, large-amplitude voltage spikes. Both phenomena are implicated in memory consolidation and learning, but the physiology and mechanisms by which they occur are not yet fully understood, see \cite{Buzsaki2006,Diekelmann2010,Niedermeyer2005,Dang2010}.

In order to study these phenomena, they anomalies must be first located and identified in the EEG data. This can be challenging because they occur for
an extremely short duration and irregularly. For example, sleep spindles and K-Complexes typically last less than 1\,s, and there are only on the order of 100 of these events over the course of an entire night. Identification of these events is further complicated by the presence of artifacts in the data, often caused by movement of the subject, but which can also occur due to electrical noise or
 loose electrodes connections. These artifacts must be ignored when attempting to identify sleep spindles and K-Complexes. Because the electric fields produced by the brain are quite weak (the induced electrical potential is on the order of 50\,$\mu$V), the signals also contain a significant noise component.

In this paper we introduce the Online Selective Anomaly Detection (OSAD)
problem which captures a particular scenario in sleep research.
As noted above, around 100 sleep spindles will occur during the course
of a night. The number of K-Complexes is much fewer. For some experiments
scientists are interested in identifying both sleep spindles  and
K-Complexes but only want to be notified with an alert when a non-spindle anomaly
occurs (for example K-Complexes).

The solution of the OSAD problem combines techniques form
both data mining and control theory. Data Mining is used
to model and  infer the normal EEG pattern per subject. Experiments
have shown that model parameters do not transfer accurately across
to other subjects. In our case we will use a Linear Dynamical
System (LDS) to model the EEG time series. Then based on
frequency analysis, we infer the sleep spindle (SS) pattern and
integrate the pattern as a disturbance into the LDS. The control theory part is used to
{\em design} a new residual which supresses SS signals but
faithfully represents other errors generated by the LDS model.
Thus by selectively supressing SS pattern, the objectives of the
OSAD problem are achieved.

For example, consider Figure~\ref{fig:example}. The top frame
shows a typical EEG time series with both the SS and KC highlighted.
The middle frame shows a typical residual time series based on
an LDS model. The bottom frame shows a new residual designed
to solve the OSAD problem. Notice that the error due to the presence
of SS is suppressed but the residual due to the appearance of
KC remains unaffected.

The main contributions of the paper are: \\
\begin{itemize}
\item We introduce the Online Selective Anomaly Detection(OSAD)
to address the requirement of selectively reporting sleep anomalies based
on specifications by domain experts.
\item In order to solve OSAD, we combine techniques from data mining
and control theory. In particular we will use a Linear Dynamic System (LDS)
to model the underlying data generating process and use control theory
techniques to design an appropriate residual system.
\end{itemize}

The rest of the paper is as follows. In Section 2, we rigorously define
the OSAD problem. In Section 3 we present our methodology to infer the
parameters of the LDS and use control theory to design a new
residual system.
 In Section 4, we apply our approach to real sleep data and
evaluate our results. We overiew related work in Section 5
and conclude in Section 6
with a summary and potential ideas for future research.

\label{section:intro}
\section{Problem Definition}
In this section we present our problem statement for selective anomaly detection.

\begin{figure*} [ht!]
\centering
\vspace{-60pt}
\includegraphics[width=0.90\textwidth]{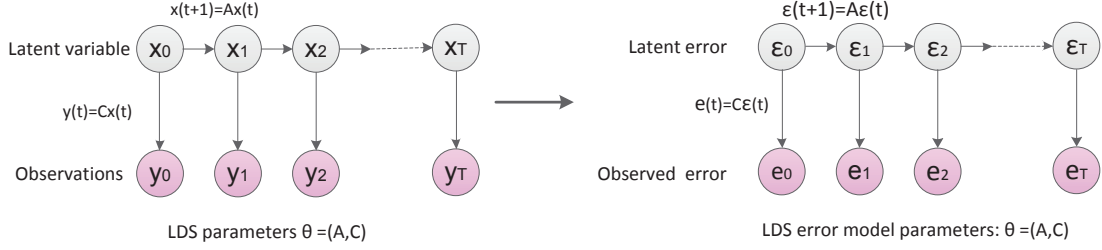}
\vspace{-70pt}
\caption{A linear dynamic system is a model which defines a linear relationship
 between the latent (or hidden) state of the model and observed outputs. The LDS parameters ${\bf A}$ and ${\bf C}$ need to be estimated from data. The LDS can
also be used to model the relationship between the latent and the observed
residuals (right figure).}
\label{fig:LDS}
\end{figure*}

The starting point  is an observed time series  of $N$ points $ y = \{y_{i}\}_{i=1}^{N}$ where
each $y_{i} \in \mathbb{R}^m $. Furthemore, we assume that the $ y$ measures
the output of a system which is generated
from a latent variable $x \in \mathbb{R}^{n}$. The relationship between
$x$ and $ y$ is governed by a standard Linear Dynamic system (LDS) model~\cite{Roweis1999} which is specified as
\[
\begin{array}{rl}
x(t+1)= &{\bf A}x(t)\\
 y(t) = & {\bf C}x(t)
\end{array}
\]
Here ${\bf A}$ is an $n \times n$  state matrix which governs the dynamics of the LDS while ${\bf C}$ is an $m \times n$ observation matrix. The modern
convention is to represent the LDS as graphical model as shown in
Figure~\ref{fig:LDS}.
The state of the system, $x$, evolves according to LDS
beginning at time $t = 0$, with value $x_0$. The standard learning problem
is as follows.
\begin{problem}[Learning Problem]
Given an observable time series $\{y_{i}\}_{i=1}^{N}$ and assuming that
the observed $ y$ and the latent $x$ are governed by an LDS,
  infer ${\bf A}$ and ${\bf C}$.
\end{problem}
The standard LDS inference problem has been extensively studied in both
the machine learning and control theory literature. Several algorithms
have been proposed including those based on gradient descent,
Expectation Maximization, subspace identification and spectral approaches
\cite{Shumway1982,Overschee1996,Ljung2001,Boots2010}. Several extensions
of LDS to include non-linear relationships as well as to include
stochastic disturbances have been proposed. However, for sleep
analysis, the above LDS will suffice. For the sake of completeness,
in the Appendix we will describe a simple but effective approach
for inferring ${\bf A}$ and ${\bf C}$  based on a spectral method
~\cite{Boots2010}.\\

The standard approach to detect outliers using an LDS is to use
the inferred ${\bf A}$ and ${\bf C}$ matrices to
compute the latent and observed error variables as:
\[
\begin{array}{rl}
\varepsilon(t) :=& x(t)-\hat{x}(t) \\
 e(t):=& y(t)-\hat{y}(t)
\end{array}
\]
where $\hat{x}$ and $\hat{y}$  are estimated using LDS. Then given a threshold parameter $\delta$, an anomaly is reported whenever,
$ e(t) > \delta$. However, our objective is not to report all
anomalies but suppress some known user-defined patterns or even known
anomalous pattern. We now formalize the notion of pattern.

\begin{definition} A {\bf pattern} ${\bf P}$ is a user-defined
matrix which operates in the latent space.
\end{definition}

In our context, we will design a specific matrix ${\bf P}$ for
a sleep spindle. The matrix ${\bf P}$ is integrated into the LDS as
\[
\begin{array}{rl}
 x(t+1)= &{\bf A}x(t)+ {\bf P}\zeta(t)\\
 y(t) = & {\bf C}x(t)
\end{array}
\]
We are now ready to define the design part of the OSAD problem.
\begin{problem}[Design Problem]
Given an LDS, a pattern ${\bf P}$ in the latent space, design
a residual $r(t)$ such that
\[
r(t)=\left\{
\begin{array}{ll}
0 & \mbox{ if }\epsilon(t) = {\bf P}\zeta(t) \\
{\bf S}e(t) & \mbox{ otherwise }
\end{array}
\right.
\]
\end{problem}

Here ${\bf S}$ is suitably defined linear transformation on $e(t)$.
Notice that the residual $r(t)$ depends both on the latent error
$\epsilon(t)$ and the observed error $e(t)$. In practice, $r(t)$ will
never be exactly zero when the pattern ${\bf P}$ is active but will
have small absolute values.

\label{section:problem}
\section{The OSAD Method}
In this section we propose a method based on statistical inference and control theory
to provide a solution of the OSAD problem. Using the LDS, we first develop a
{\tt Dynamic Residue Model} (DRM). Then we will show how to adjust the DRM parameters in order to design
a residual $r(t)$ which will satisfy the
constraints of the problem, i.e. the selected anomalous pattern will be canceled (or projected out) in the
generated residual space.

\subsection{DRM Formulation}
Assume data is generated by an LDS. Any deviation of the state from its expected
value can be captured by a structured error model. Intuitively, the discrepancy between the observed error $e(t)$ and latent error $\varepsilon(t)$
is modeled by the same LDS (because of linearity):
\[
\begin{array}{rl}
\varepsilon(t+1)=& {\bf A}\varepsilon(t)+ {\bf P} \xi(t)\\
            e(t)=& {\bf C}\varepsilon(t)
\end{array}
\]

The above  error model  can be used to detect changes occurring in the
latent space.\\

We design a feedback loop (as shown in Figure \ref{fig:Feedback}) to effect the output of
the error model. In particular a function of the residual will be used
to manipulate the changes in the error. The design objective will be
to map the anomalies generated by the $\bf{P}$ pattern into the
null space of the new residual. The DRM based on this feedback
design is developed as follows:
\begin{figure*}
\centering
\vspace{-30pt}
\hspace{-18pt}
\includegraphics[width=0.66\textwidth]{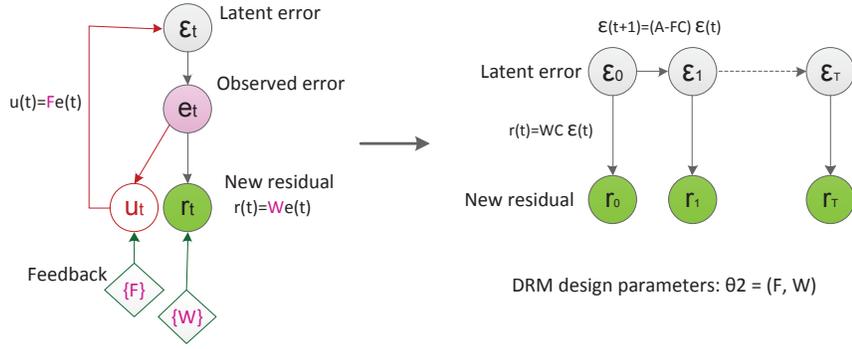}
\vspace{-38pt}
\caption{Using parameter {\bf F} a virtual input $u(t)$ is generated to feed the error back to the latent space. The error $e(t)$ is is then calibrated by {\bf W} to generate a new residual space $r(t)$.}
\label{fig:Feedback}
\end{figure*}

To design the feedback we define two transformation matrices ${\bf W}$ and ${\bf F}$ for error values to be weighted as:
\[
\begin{array}{rl}
r(t):=& {\bf W}e(t) \\
u(t):=&{\bf F}e(t)
\end{array}
\]
${\bf F}$ will be used as the feedback gain matrix and maps the error to the feedback vector $u(t)$, and ${\bf W}$ is the residual weighting matrix that generates the new residual $r(t)$.
Now feeding back $u(t)$ into the LDS (as shown in  Figure~\ref{fig:LDS}), with $u(t),$ the residual dynamic model will be:
\[
\begin{array}{rl}
                 \hat{x}(t + 1) =& {\bf A}\hat{x}(t) + u(t)   \\
                                =& {\bf A}\hat{x}(t) +{\bf F}e(t)\\
                                =& {\bf A}\hat{x}(t) +{\bf F}(y(t) - \hat{y}(t))\\
                                =& {\bf A}\hat{x}(t) +{\bf F}({\bf C}x(t) - {\bf C}\hat{x}(t))\\
                                =& {\bf A}\hat{x}(t) +{\bf F}{\bf C}x(t) - {\bf F}{\bf C}\hat{x}(t)\\
                                =& ({\bf A}-{\bf F}{\bf C})\hat{x}(t) + {\bf F}{\bf C}x(t)\\
                                =& ({\bf A}-{\bf F}{\bf C})\hat{x}(t) + {\bf F} y(t)\\
\end{array}
\]

Notice that since the residual is a linear transformation of the error, its
rank (suppose $r(t) \in \mathbb{R}^p$) can not be larger than the observation dimension, i.e., $p\leq m$.

We are now able  to define the dynamic of the latent error as:
\[
\begin{array}{rl}
                \varepsilon(t+1) =&x(t+1)-\hat{x}(t+1) \\
                                 =&{\bf A}x(t)-({\bf A}-{\bf F}{\bf C})\hat{x}(t)- {\bf F} y(t) \\
                                 =&{\bf A}x(t)-{\bf A}\hat{x}(t)-{\bf F}{\bf C})\hat{x}(t)+ {\bf F}{\bf C}x(t) \\
                                 =&({\bf A}-{\bf F}{\bf C})(x(t)-\hat{x}(t)) \\
                                 =&({\bf A}-{\bf F}{\bf C})\varepsilon(t)\\
\end{array}
\]
and the residue $r(t)$ is obtained as:
\[
\begin{array}{rl}
                    r(t)=& {\bf W}(y(t) - \hat{y}(t))\\
                        =& {\bf W}({\bf C}x(t) - {\bf C}\hat{x}(t))\\
                        =& {\bf W}{\bf C}(x(t) - \hat{x}(t))\\
                        =& {\bf WC} \varepsilon(t)
\end{array}
\]
We therefore have the following  dynamic model for the latent error:
\[
\begin{array}{rl}
                \varepsilon(t+1) =&({\bf A}-{\bf F}{\bf C})\varepsilon(t)\\
                            r(t) =& {\bf WC} \varepsilon(t)

\end{array}
\]
Notice that the observed residue $r(t)$ is governed by state error $\varepsilon(t)$ through matrix ${\bf WC}$ while it evolves in time through ${\bf A}-{\bf F}{\bf C}$.

To simplify the notation, denote ${\bf C}_f={\bf WC}$ and ${\bf A}_f={\bf A}-{\bf FC}$.
The DRM is then defined as:
\[
\begin{array}{rl}
                \varepsilon(t+1) =&{\bf A}_f \varepsilon(t)\\
                            r(t) =& {\bf C}_f \varepsilon(t)

\end{array}
\]
The graphical diagram for this error model is shown in Figure~\ref{fig:Feedback}.

\subsection{OSAD Parameter Design}
\begin{figure}
\hspace{-20pt}
\includegraphics[width=0.85\textwidth]{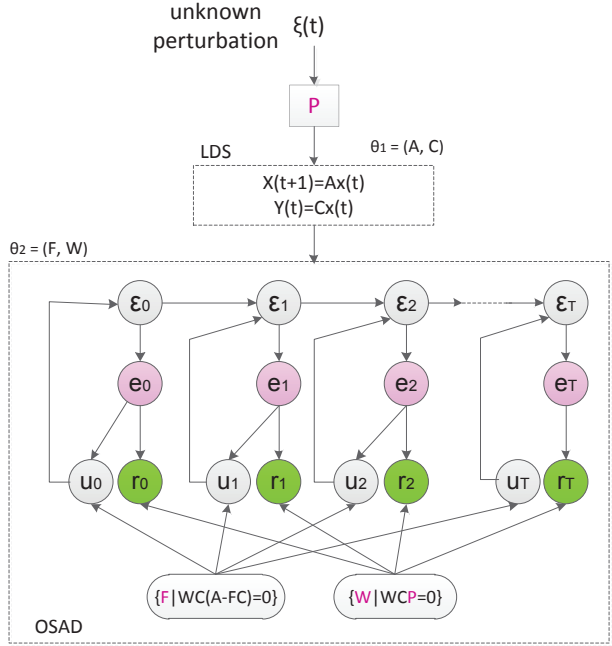}
\hspace{-40pt}
\caption{The complete diagram of OSAD. Using parameters {\bf W} and {\bf F} the residue space $r(t)$ is calibrated to cancel the impact of ${\bf P}\xi(t)$.}
\label{fig:Filter}
\end{figure}

In this section we address the problem of
designing the ${\bf F}$ and ${\bf W}$ matrix with objective of making the
DRM insensitive to anomalies generated by ${\bf P}$. The overarching
design is shown in Figure~\ref{fig:Filter} and is related to the use of control theory for fault diagnosis \cite{Patton1997,Patton2000,Chen1999}. A typical LDS model will
output the observed error $e(t)$. However, the OSAD model has a feedback
loop which takes ${\bf W}$ and ${\bf F}$ matrices as input and return a variable
$u(t)$ which is fed back into the model. The observed error is also transformed
by a ${\bf W}$ matrix. The ${\bf F}$ and the ${\bf W}$ matrices satisfy the constraints
which involve the ${\bf A}$, ${\bf C}$ and the ${\bf P}$ matrices.

Since the model is time-dependent, we follow a standard approach and
map the model into the frequency domain using a  $\mathcal{Z}$-transform to design the ${\bf W}$ and ${\bf F}$ matrices. In the frequency domain, it will be easier
to design matrices ${\bf W}$ and ${\bf F}$ such that ${\bf WC}({\bf A-FC}) = 0$ and
${\bf WCP} =0$.

\begin{definition} The Z-transform of a discrete-time sequence $x(k)$ is the series $X(z)$ defined as
\[
X(z)= \mathcal{Z}\{x(k)\}=\sum _{0}^{\infty}x(k)z^{-k}.
\]
\end{definition}
\begin{observation} Two important (and well known)  properties of the
 Z-transform are linearity and time shifting:
\[
ax(k)+by(t) \overset{Z}{\longleftrightarrow} aX(z)+ b Y(Z)
\]
\[
x(k+b) \overset{Z}{\longleftrightarrow} z^{b}X(z)
\]
\end{observation}

Applying Z-transform $\mathcal{Z}()$ to the DRM yields:

\[
\begin{array}{rl}
                  z\mathcal{E}(z) =&{\bf A}_f\mathcal{E}(z)+ {\bf P} \boldsymbol\xi(z) \\
                  \mathcal{E}(z)=&(z{\bf I}-{\bf A}_f)^{-1}{\bf P}\boldsymbol\xi(z) \\
                    \end{array}
\]
and:
\[
\begin{array}{rl}
                          R(z) =& {\bf C}_f \mathcal{E}(z)\\
                              =&[ {\bf C}_f(z{\bf I}-{\bf A}_f)^{-1}{\bf P}] \boldsymbol\xi(z)
\end{array}
\]
in which $\boldsymbol\xi(z)=\mathcal{Z}(\xi(t))$, $\boldsymbol\vartheta=\mathcal{Z}(\vartheta(t))$, $R(z)=\mathcal{Z}(r(t))$.
The transfer gain between $\boldsymbol\xi$ and $R$:
\[
\begin{array}{rl}
                 {\bf G}_{\xi}(z) :=   & {\bf C}_f(z{\bf I}-{\bf A}_f)^{-1}{\bf P}
\end{array}
\]
Thus if ${\bf G}_{\xi}$ would be zero, the residual $R(z)$ is independent of the $\boldsymbol\xi(z)$. In the other word, to make $R(z)$ independent of $\boldsymbol\xi(z)$, one must null the space of $ {\bf G}_{\xi}(z) $. Then whenever $\mathcal{P}$ occurs it is transferred by a zero gain to the residual space. To find the null space ${\bf G}_{\xi}(z)=0$, we expand it as:
\[
\begin{array}{rl}
                        {\bf G}_{\xi}(z) = & z^{-1}{\bf C}_f({\bf I+A}_fz^{-1} + {\bf A}_f^2z^{-2}+...){\bf P}\\
                                        = & 0
\end{array}
\]

The sufficient conditions for ${\bf G}_{\xi}(z)$ to be nulled are ${\bf C}_f{\bf P} = 0$ and either ${\bf C}_f{\bf A}_f = 0$ or ${\bf A}_f {\bf P} = 0$.
Thus we have the following result.\\

\begin{theorem} For a DRM, a sufficient condition for ${\bf G}_{\xi}(z) = 0$ is

\[
{\bf C}_f{\bf P}  = 0 \mbox{ and }
\{{\bf C}_f{\bf A}_f = 0 \mbox{ or } {\bf A}_f {\bf P} = 0\}
\]

\end{theorem}

Now as ${\bf C}_f= {\bf WC}$,  for ${\bf C}_f{\bf P} = 0$ it is sufficient
that ${\bf WC}$ be orthogonal to ${\bf P}$.  Furthermore for
${\bf C}_f{\bf A}_f = 0$, it
is sufficient to design a matrix ${\bf A}_f$ such that its left
eigevectors corresponding to the zero eigenvalue are orthogonal to ${\bf P}$.
Similarly, for ${\bf A}_f {\bf P} = 0$, it is sufficient to
design a matrix ${\bf A}_f$, such that the right eigenvectors corresponding
to the zero eigenvalues are orthogonal to ${\bf P}$. See Appendix~\ref{app:proof}.
%

Now, it design a system which operates in an online fashion we proceed as follows.
From the definition of residue:
\[
\begin{array}{rl}
                r(t) =& {\bf W}[y(t)-\hat{y}(t)]
\end{array}
\]
\noindent
Using the Z-transform, the computational form of the residual will be:
\[
\begin{array}{rl}
                R(z) =&[{\bf W}-{\bf C}_f(z{\bf I-A}_f)^{-1}{\bf F}]Y(z)

\end{array}
\]
\noindent
Since ${\bf C}_f{\bf A}_f =0$:
\[
\begin{array}{rl}
                 {\bf C}_f(z{\bf I-A}_f)^{-1}{\bf F}= &z^{-1}{\bf C}_f
\end{array}
\]
\noindent
Replacing this result to the above $R(z)$ equation:
\[
\begin{array}{rl}
                R(z) =&({\bf W}-z^{-1}{\bf C}_f{\bf F})Y(z)
\end{array}
\]
\noindent
Applying the inverse Z-transform, the equation will be:
\[
\begin{array}{rl}
                r(t) =&\begin{bmatrix} {\bf W} \quad -{\bf C}_f{\bf F} \end{bmatrix} \begin{bmatrix} y(t)\\ y(t-1) \end{bmatrix}

\end{array}
\]
\noindent
This clearly says that the residual can be represented directly in terms of the
observations. This property is crucial to make the anomaly detection system operate in near real-time.

\subsection{Eigenpair Assignment and the ${\bf F}$ Matrix}
In this section we explain the eigenpair assignment problem and its solution which is used for designing the
matrix ${\bf F}$. Recall from Theorem 1, that we require either ${\bf C}_{f}{\bf A}_{f} =0$ or
${\bf A}_{f}{\bf P} = 0$.

\begin{problem}
Given a set of scalars $\{\lambda_i\}$ and a set of n-vectors $\{v_i\}$ (for $ i=1,2,...,n$), find a real matrix ${\bf A}_o$ ($m \times n$) such that the eigenvalues of ${\bf A}_o$ are precisely those of the set of scalars $\{\lambda_i\}$ with corresponding eigenvectors the set $\{v_i\}$.
\end{problem}
Given the residue model transition matrix ${\bf A}_f={\bf A}-{\bf FC}$, the problem is to find a matrix ${\bf F}$
such that this matrix has the eigenvalues $\{\lambda_i\}$ corresponding to eigenvectors $\{v_i\}$,i.e.,:
\[
( {\bf  A-FC} ) v_i = \lambda_i v_i
\]

or:
\[
\begin{bmatrix}{\bf A}-\lambda_i{\bf I}& &{\bf C'} \end{bmatrix} \begin{bmatrix} v_i \\ -{\bf F} v_i \end{bmatrix} = 0
\]
\noindent
Define $q_i:=-{\bf F} v_i$, then:
\[
\begin{bmatrix} {\bf A}-\lambda_i{\bf I} & &{\bf C'}\end{bmatrix} \begin{bmatrix} v_i \\ q_i \end{bmatrix} = 0
\]
\noindent
The implication of the above statement is of great importance: The vectors $\begin{bmatrix} v_i & q_i \end{bmatrix}'$ must be in the \emph{kernel space} of $\begin{bmatrix} {\bf A}-\lambda_i{\bf I} & &{\bf C'}\end{bmatrix}$, meaning, for $i=1,2,...,n$:
\[
\begin{bmatrix} q_1 & q_2& ...& q_n \end{bmatrix} = \begin{bmatrix} -{\bf F}v_1&  -{\bf F}v_2&...& -{\bf F}v_n \end{bmatrix}
\]
\noindent
The matrix ${\bf F}$ now can be obtained as:
\[
{\bf F}=-\begin{bmatrix} q_1 & q_2& ...& q_n \end{bmatrix} \begin{bmatrix} v_1&  v_2&...& v_n \end{bmatrix}^{+}
\]

where '+' stands for pseudoinverse. The whole procedure is summarized in Algorithm~\ref{alg:EA}.

\begin{algorithm} [!t]
\caption{Find {\bf F} such that the set $\{\lambda_i, v_i\}$ be the eigenpairs of ${\bf A}-{\bf FC}$}

\begin{algorithmic} [1]
\vspace{4pt}
\STATE Input {\bf A,C}, $\lambda_{i} = 0$ $\forall i$ and $v_{i} = P(:,i)$.
\vspace{4pt}
\STATE Output ${\bf F}$ such that ${\bf (A - FC)P = 0}$. \\
\vspace{4pt}
\FOR {$i=1:n$}
\vspace{3pt}
\STATE $\phi_i = \textit{null} \begin{bmatrix} {\bf A}-\lambda_i{\bf I} & {\bf
C'}\end{bmatrix}$ 
\vspace{3pt}
\STATE Find an element $[v_{i} \mbox{  }q_{i}]' \in \phi_i$
\vspace{3pt}
\ENDFOR
\vspace{4pt}
\STATE  ${\bf F}=-\begin{pmatrix} q_1 & q_2& ...& q_n \end{pmatrix} \begin{pmatrix} v_1&  v_2&...& v_n \end{pmatrix}^{+}$
\end{algorithmic}

\label{alg:EA}
\end{algorithm}

\subsection{Degrees of Freedom of ${\bf P}$}
There is an an important constraint that the matrix ${\bf P}$ must satisfy
for the DRM approach to be valid solution of the OSAD problem.
As the ${\bf WCP} = 0$, a necessary condition is that
\[
                        \mbox{rank}({\bf P})\leq \mbox{rank}({\bf C})
\]
In the other word, the effective number of independent perturbations generated
by the matrix ${\bf P}$ is bounded by the effective number of
independent measurements governed by the observation matrix
${\bf C}$, see~\cite{Patton2000}. For example, if ${\bf C}$ is the independent
matrix on an LDS where the state vector has dimensionality $n$, then the
rank of the {\bf P} matrix must be less than $(n-1)$.

\subsection{Inferring the Matrix ${\bf P}$}
The OSAD model is predicated on the existence of a ${\bf P}$ matrix. This
matrix can be provided by a domain expert or can sometimes be inferred
from data. For example, in the case of sleep spindle, frequency analysis
shows that sleep spindles occur in the interval twelve to fourteen Hz.
The exact frequency can change from one subject to another. The signature
for K-Complexes is more a function of the amplitude of the signal rather
than the frequency.

We now show how to construct a ${\bf P}$ matrix from data. For example,
suppose there exists a frequency/peridicity $\mathcal{T}=f^{-1}$ in the EEG time series or:
\[
\begin{array}{rl}
x(t+\mathcal{T})&=x(t)
\end{array}
\]
Replace this in linear dynamics:

\[
\begin{array}{rl}
x(t+1)&={\bf A}x(t)\\
      &={\bf A}x(t+\mathcal{T})
\end{array}
\]
Applying z-transform:
\[
\begin{array}{rl}
zX(z)&={\bf A}z^\mathcal{T}X(z)
\end{array}
\]
Using Tailor expansion we expand $z^\mathcal{T}$ around $z=1$:
\[
z^\mathcal{T}\approx 1+ \alpha + \beta z + \gamma z^2
\]
where $\alpha  = 0.5 \mathcal{T}(\mathcal{T}-3)$, $\beta  = 0.5\mathcal{T}(\mathcal{T}-1) $ and $\gamma  = - \mathcal{T}(\mathcal{T}-2)$. An approximation by this expansion will be:
\[
\begin{array}{rl}
zX(z)&\approx {\bf A}X(z) + \alpha {\bf A}X(z)+ \beta z {\bf A}X(z) + \gamma z^2 X(z)
\end{array}
\]
Returning to the time-domain, we obtain
\[
\resizebox{\columnwidth}{!}{$
\begin{array}{rl}
x(t+1) &\approx {\bf A}x(t) + \alpha {\bf A} x(t) + \beta {\bf A} x(t+1) + \gamma {\bf A} x(t+2)\\
 &\approx {\bf A}x(t) + [ \alpha {\bf A}  \quad \beta {\bf A}  \quad \gamma {\bf A}] [x(t)\quad x(t+1)\quad x(t+2)]'
\end{array}
$
}
\]
\subsection{Summary Example}
To summarize, the solution of the OSAD problem requires the availability of the following
matrices:
\begin{table} [H]
\caption{Parameters for learning and design}
\resizebox{.99\hsize}{!}{
\begin{tabularx} {0.57\textwidth}{ >{\bfseries \sffamily}c  p{4pt}  >{\sffamily}l p{4pt} >{\sffamily}l  }
\toprule
\addlinespace[12pt]
        Matrix  &    &  Description & &  Source  \\ [0.99ex]
                   \cmidrule{3-3}  \cmidrule{5-5}   \\
 {\bf A} &    & The State Matrix &    & Inferred from data\\ [0.99ex]
 {\bf C} &    & The Observation Matrix &   & Inferred from data   \\ [0.99ex]
 {\bf P} &    & The Pattern  Matrix&     & Given by domain-expert \\ [0.99ex]
 {\bf F} &    & Feedback Gain Matrix &    & Designed using Theorem 1      \\ [0.99ex]
{\bf W} &    & Error Weighting  Matrix &  & Designed using Theorem 1       \\ [0.99ex]
\bottomrule
\end{tabularx}
}
\label{table:res}
\end{table}

We will now give a concrete example. Assume we have an LDS system given as

\[
\begin{array}{rl}
\varepsilon(t+1)=& {\bf A}\varepsilon(t)+ {\bf P} \xi(t)\\
            e(t)=& {\bf C}\varepsilon(t)
\end{array}
\]

Assume have identified the ${\bf A}$ and ${\bf C}$ matrices as
\[
{\bf A}=
\begin{pmatrix}
0.5 & 0.3 \\
0.3 & 0.2
\end{pmatrix}
\mbox{ and }
{\bf C}=
\begin{pmatrix}
1 &  0 \\
0 & 1
\end{pmatrix}
\mbox{ and }
{\bf P}=
\begin{pmatrix}
1 & 1 \\
2 & 2
\end{pmatrix}
\]
Now, to form the OSAD model, we have to identify ${\bf W}$ and ${\bf F}$
such that:
\begin{enumerate}
\item
 ${\bf W}$ is in the null space of ${\bf CP}$ and
\item
${\bf A - FC}$
has its left eigenvectors (corresponding to the ${\bf 0}$ eigenvalue ),
the rows of ${\bf WC}$.
\end{enumerate}
Since ${\bf C}$ is the identity matrix,  an example of ${\bf W}$  is
\[
{\bf W} =
\begin{pmatrix}
2 & - 1 \\
2 & -1
\end{pmatrix}
\]
Similarly, an example of ${\bf F}$ matrix is
\[
{\bf F} =
\begin{pmatrix}
0.0 & 0.2 \\
-0.7 & 0
\end{pmatrix}
\]
As mentioned, the residual matrix is given by
\[
r(t) =
\begin{pmatrix}
1.3 & -1.4 \\
1.3 & -1.4
\end{pmatrix}
\begin{pmatrix}
y(t) \\
y(t-1)
\end{pmatrix}
\]

\label{section:solution}
\section{Experimental Result}
We now report on the experiments that have been carried out to test the effective of the proposed
OSAD solution on sleep data. Our particular focus will be determining if OSAD can recognize
sleep spindle and K-Complex anomalies and selectively raise an alert for non-Spindle anomalies.

\subsection{Sleep Data Set}
\begin{figure}[!t]
\centering
        \includegraphics[width=0.5\textwidth]{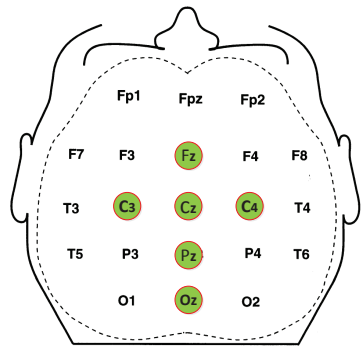}
        \caption{The position of scalp electrodes for EEG experiment follows the International 10-20 system \cite{Niedermeyer2005,Buzsaki2006}.}
\label{fig:Sensors}
\end{figure}

Our data set consists of EEG time series from four health controls (age 25-36) as described in \cite{DRozario2013}.
Recordings were made with an Alice-4 system (Respironics, Murraysville PA, USA) at the Woolcock Institute of Medical Research, at Sydney University, using 6 EEG channels with a sampling rate of 200\,Hz, and electrodes positioned according to the International 10-20 system \cite{Niedermeyer2005,Buzsaki2006}, see Figure~\ref{fig:Sensors}. In this study we only examine the Cz electrode.
A notch filter at 50\,Hz (as provided by the Alice-4 system) was used to remove mains voltage interference. No other hardware filters were used. Spindles and K-Complexes were labeled using another automation program and then
manually evaluated. As previously noted, while data from only four subjects were used, a typical EEG
session generates a large amount of personal data.

\subsection{Inference of ${\bf A}$ and ${\bf C}$ Matrices}
\begin{figure}
        \centering
        \begin{subfigure}[b]{0.22\textwidth}
                \includegraphics[width=\textwidth]{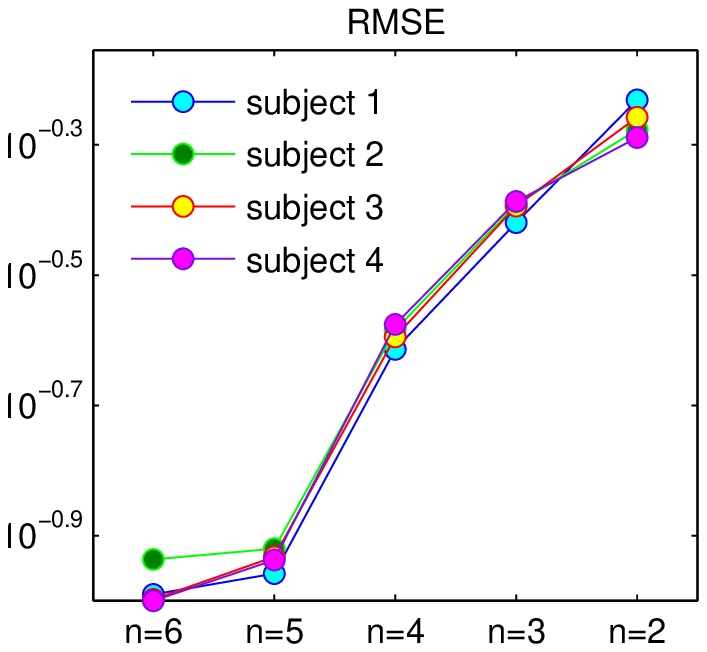}
                \caption{Subspace method}
                \label{fig:subspace}
        \end{subfigure}~
       \begin{subfigure}[b]{0.22\textwidth}
        \includegraphics[width=\textwidth]{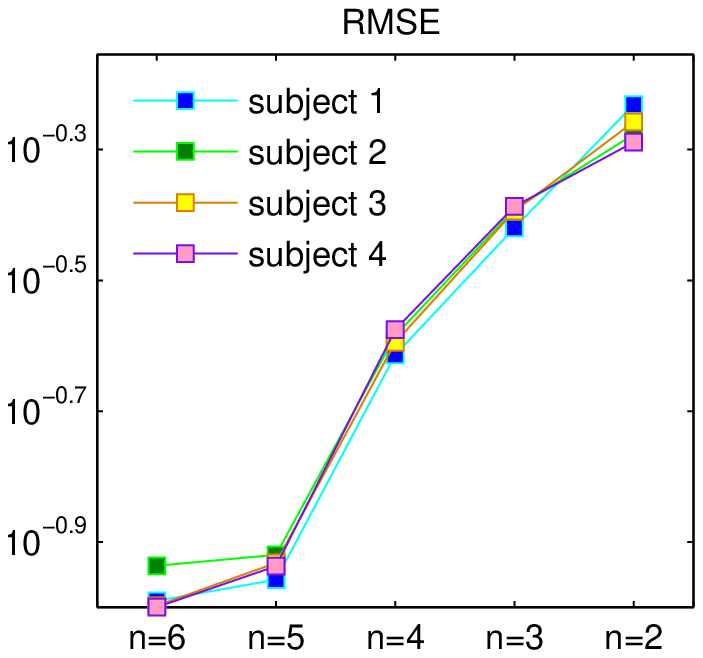}
        \caption{Spectral method}
         \label{fig:spectral}
        \end{subfigure}
        \caption{The RMSE error obtained from both methods are comparable. Notice the RMSE increases as the rank of LDS is reduced.}
\label{fig:time}
\end{figure}

Our first task is to learn the ${\bf A}$ and ${\bf C}$ matrices from the LDS for each subject. Others have
reported, and our experiments confirm, that EEG of each subject tends to different and separate models
need to learnt per subject. For each subject we took a sample of size 2000 (10 seconds) of EEG time series
which did not contain either sleep spindle or K-Complex. We then formed a $2000 \times 6$ data
matrix, ${\bf O}$. The columns of the ${\bf O}$ matrix are time series associated
with the six channels of EEG. We used both subspace and spectral methods to infer the matrices
${\bf A}$ and ${\bf C}$. Both these methods are based on SVD decomposition of the ${\bf O}$ matrix
and require as input the rank required of the inferred matrices. We evaluated the inferred matrices using RMSE and the results are shown
in Figure~\ref{fig:subspace} and Figure~\ref{fig:spectral}. Both the subspace and spectral
methods have similar performance and RMSE goes up significantly when the rank falls below five.
We selected a rank six matrix (maximum possible rank) for both ${\bf A}$ and ${\bf C}$. In terms
of running time, the two methods are comparable as we have to carry out an SVD of a relatively
small $6 \times 6$ matrix.

\subsection{Detection of SS  and K-Complex}
For each of the four subjects, statistics of the labeled sleep spindles and K-Complexes
and those detected by the LDS are shown in Table~\ref{table:res1}. For LDS detection, we used a threshold
derived from CUSUM which automatically adjusts for mean and standard deviation of the observed
residual time series $e(t)$. To specify a CUSUM threshold we applied the alpha and beta approach in \cite{montgomery2004} and we set the probabilities of a false positive and a false negative to $10^{-4}$ and the change detection parameter to 1 sigma, in all subjects. \\

In all four subjects, the LDS residual slightly under predicts the number of spindles and K-Complexes.
Since each labeled and predicted SS and K-Complex spans a time-interval, we have modified
the definitions of {\tt precision} and {\tt recall} to account for the intervals. For a
given subject, let $\{[a_{i},b_{i}]\}_{i=1}^{n}$ be the intervals of the labeled anomalies
(spindles or K-Complex).
Let $\{[a^{'}_{j},b^{'}_{j}]\}_{j=1}^{m}$ be the predicted spindles. Then

\[
{\tt precision} = \frac{\sum_{i=1}^{n}\sum_{j=1}^{m}|[a_{i},b_{i}] \cap [a^{'}_{j},b^{'}_{j}]|}{\sum_{j=1}^{m}|[a^{'}_{j},b^{'}_{j}]|}
\]

and

\[
{\tt recall} = \frac{\sum_{i=1}^{n}\sum_{j=1}^{m}|[a_{i},b_{i}] \cap [a^{'}_{j},b^{'}_{j}]|}{\sum_{j=1}^{m}|[a_{j},b_{j}]|}
\]

Here,  $|[a_{i},b_{i}]|$, is the number of points in the time interval $[a_{i},b_{i}]$. With these
definitions in place, Table~\ref{table:res2} and Table~\ref{table:res3} show the precision and recall SS and K-Complex across
alls the subjects. In general both precision and recall are high across subjects, but precision
is significantly more higher than recall. For SS, the recall varies more than precision ranging
for 71.24\% to 97.18\%. Also notice that the length of detection of both SS and K-Complex is
higher compared to their labeled lengths.

\begin{table} [!t]
\caption{Summary statistics of results. LDS is quite accurate but tends to over-predict the number
of anomalies.}
\centering
\vspace{-5pt}
\resizebox{.99\hsize}{!}{
\begin{tabularx} {0.62\textwidth}{ >{\bfseries \sffamily}r  p{5pt}  >{\sffamily}c  >{\sffamily}c  p{2pt} >{\sffamily}c   >{\sffamily}c}
\toprule
\addlinespace[12pt]
          &    &  \multicolumn{2}{c}{No. of Labeled Anomalies }  &  &  \multicolumn{2}{c}{ No. of Detected Anomalies}\\
\addlinespace[-2pt]
                      \cmidrule{3-4}                                               \cmidrule{6-7}       \\
\addlinespace[-10pt]
          &    &  Spindle &  K-Complex  &  &Spindle &  K-Complex \\ [0.99ex]
\midrule
\addlinespace[3pt]
subject 1 &    &  170&  277  &  &164 &  251 \\ [0.99ex]
subject 2 &    &  6 &  13  &  &6 &  11 \\ [0.99ex]
subject 3 &    &  23 &  38  &  &21 &  37 \\ [0.99ex]
subject 4 &    &  141 &  205  &  &132 &  186 \\ [0.99ex]
\bottomrule
\end{tabularx}
}
\label{table:res1}
\end{table}

\begin{table} [!t]
\caption{Summary statistics for spindles. LDS has higher precision than recall and total length
of predicted interval is higher than the length of labeled intervals.}
\centering
\vspace{-5pt}
\resizebox{.99\hsize}{!}{
\begin{tabularx} {0.6\textwidth}{ >{\bfseries \sffamily}r  p{5pt}  >{\sffamily}c  >{\sffamily}c  p{2pt} >{\sffamily}c   >{\sffamily}c}
\toprule
\addlinespace[12pt]
          &    &  \multicolumn{2}{c}{Total time of spindles}  &  &  \multicolumn{2}{c}{Performance}\\
                      \cmidrule{3-4}                                               \cmidrule{6-7}       \\
 \addlinespace[-10pt]
          &    &  Labeled in min &  Detected in min &  & Recall &  Precision \\ [0.99ex]
\midrule
\addlinespace[2pt]
subject 1 &    &  129.8&  168.74  &  &71.24\% &  95.53\% \\ [0.99ex]
subject 2 &    &  3.45 &  3.55  &  &97.18\% &  97.38\% \\ [0.99ex]
subject 3 &    &  15.15 &  16.23  &  &83.88\% &  95.66\% \\ [0.99ex]
subject 4 &    &  93.5 &  103.2  &  &79.15\% &  95.42\% \\ [0.99ex]
\bottomrule
\end{tabularx}
}
\label{table:res2}
\end{table}

\begin{table} [!t]
\caption{Summary statistics for K-Complex. Both precision and recall are high. Total length
of predicted interval is higher than labeled intervals.}
\centering
\vspace{-5pt}
\resizebox{.99\hsize}{!}{
\begin{tabularx} {0.6\textwidth}{ >{\bfseries \sffamily}r  p{5pt}  >{\sffamily}c  >{\sffamily}c  p{2pt} >{\sffamily}c   >{\sffamily}c}
\toprule
\addlinespace[12pt]
          &    &  \multicolumn{2}{c}{Total time of K-Complex}  &  &  \multicolumn{2}{c}{Performance}\\
                      \cmidrule{3-4}                                               \cmidrule{6-7}       \\
                      \addlinespace[-10pt]
          &    &  Labeled in min &  Detected in min &  & Recall &  Precision \\ [0.99ex]
\midrule
\addlinespace[2pt]
subject 1 &    &  198.23&  216.35  &  &90.45\% &  93.43\% \\ [0.99ex]
subject 2 &    &  11.48 &  11.25  &  &92.76\% &  94.12\% \\ [0.99ex]
subject 3 &    &  21.39 &  24.56  &  &91.01\% &  92.06\% \\ [0.99ex]
subject 4 &    &  147.68 &  160.49  &  &91.28\% &  93.73\% \\ [0.99ex]
\bottomrule
\end{tabularx}
}
\label{table:res3}
\end{table}
\subsection{Evaluation across Subjects}
We now investigate the transfer properties of the inferred LDS across subjects. That is, we learn
the ${\bf A}$ an ${\bf C}$ matrices on one subject and evaluate it against an another. We just
focus on the anomaly. The {\tt recall} and {\tt precision} results are shown in Table~\ref{table:res4} and Table ~\ref{table:res5} respectively. The
diagonal of the table corresponds to the results in Table~\ref{table:res2} and Table~\ref{table:res3}. It is clear that there is
a substantial reduction in accuracy and that indeed the EEG of subjects varies substantially. We
have also computed the "average" ${\bf A}$ and ${\bf C}$ matrix and evaluated against all
the four subjects. The results are shown in Table~\ref{table:res6}. While there is an improvement compared
to results in Table~\ref{table:res4} and Table~\ref{table:res5}, the absolute performance is still quite low compared to the situation
where the learning was customized per individual subject.

\begin{table} [!t]
\caption{Recall across subjects. A substantial reduction in accuracy when model of one
subject is evaluated against the EEG of another.}
\centering
\vspace{-5pt}
\resizebox{.99\hsize}{!}{
\begin{tabularx} {0.72\textwidth}{ >{\bfseries \sffamily}r  p{4pt}  >{\sffamily}c p{1pt} >{\sffamily}c p{1pt} >{\sffamily}c  p{1pt} >{\sffamily}c }
\toprule
\addlinespace[12pt]
          &    &  $A_1,C_1,W_1,F_1$ & &  $A_2,C_2,W_2,F_2$ &  &$A_3,C_3,W_3,F_3$ & & $A_4,C_4,W_4,F_4$ \\ [0.99ex]
                   \cmidrule{3-3}  \cmidrule{5-5} \cmidrule{7-7} \cmidrule{9-9} \\
\addlinespace[2pt]
subject 1 &    &  71.24\%& & 38.13\%  &  &44.13\% & & 41.29\% \\ [0.99ex]
subject 2 &    &  41.32\% &  &97.18\%  &  &35.26\% & & 37.85\% \\ [0.99ex]
subject 3 &    &  48.74\% & & 43.21\% &  &83.88\% & & 44.43\% \\ [0.99ex]
subject 4 &    &  51.26\% & & 43.81\%  &  &35.36\% &  &79.15\% \\ [0.99ex]
\bottomrule
\end{tabularx}
}
\label{table:res4}
\end{table}

\begin{table} [!t]
\caption{Precision across the subjects. Again, a substantial reduction in accuracy when
model of one subjected is evaluated against another.}
\centering
\vspace{-5pt}
\resizebox{.99\hsize}{!}{
\begin{tabularx} {0.72\textwidth}{ >{\bfseries \sffamily}r  p{4pt}  >{\sffamily}c p{1pt} >{\sffamily}c p{1pt} >{\sffamily}c  p{1pt} >{\sffamily}c }
\toprule
\addlinespace[12pt]
          &    &  $A_1,C_1,W_1,F_1$ & &  $A_2,C_2,W_2,F_2$ &  &$A_3,C_3,W_3,F_3$ & & $A_4,C_4,W_4,F_4$ \\ [0.99ex]
                   \cmidrule{3-3}  \cmidrule{5-5} \cmidrule{7-7} \cmidrule{9-9} \\
\addlinespace[12pt]
subject 1 &    &  95.53\%& & 41.11\%  &  &47.19\% & & 43.67\% \\ [0.99ex]
subject 2 &    &  39.54\% &  &97.38\%  &  &37.82\% & & 39.21\% \\ [0.99ex]
subject 3 &    &  48.21\% & & 41.29\% &  &95.77\% & &41.83\% \\ [0.99ex]
subject 4 &    &  51.77\% & & 53.34\%  &  &33.49\% &  &95.42\% \\ [0.99ex]
\bottomrule
\end{tabularx}
}
\label{table:res5}
\end{table}

\begin{table} [!t]
\caption{Recall and Precision on each subject evaluated against an averaged model. Again,
a substantial reduction in accuracy compared to individual models.}
\centering
\vspace{-5pt}
\resizebox{.7\hsize}{!}{
\begin{tabularx} {0.38\textwidth}{ >{\bfseries \sffamily}r  p{6pt}  >{\sffamily}c p{6pt} >{\sffamily}c }
\toprule
\addlinespace[12pt]
          &    &  $Recall$ & &  $Precision$ \\ [0.99ex]
                   \cmidrule{3-3}  \cmidrule{5-5} 
\addlinespace[2pt]
subject 1 &    &  69.35\% &  &51.39\% \\ [0.99ex]
subject 2 &    &  65.43\% &  &57.22\%  \\ [0.99ex]
subject 3 &    &  61.77\% &  &61.47\%  \\ [0.99ex]
subject 4 &    &  68.12\% &  &53.92\%  \\ [0.99ex]
\bottomrule
\end{tabularx}
}
\label{table:res6}
\end{table}
\subsection{Performance of Designed Residual}
\begin{figure}
        \centering
        \begin{subfigure}[b]{0.22\textwidth}
                \includegraphics[width=\textwidth]{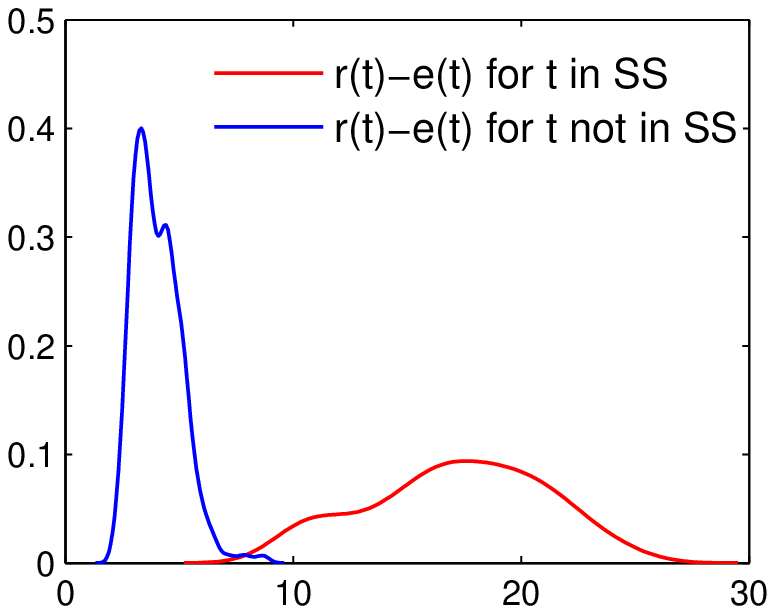}
                \caption{subject 1}
                \label{fig:DoS hist}
        \end{subfigure}~
       \begin{subfigure}[b]{0.22\textwidth}
        \includegraphics[width=\textwidth]{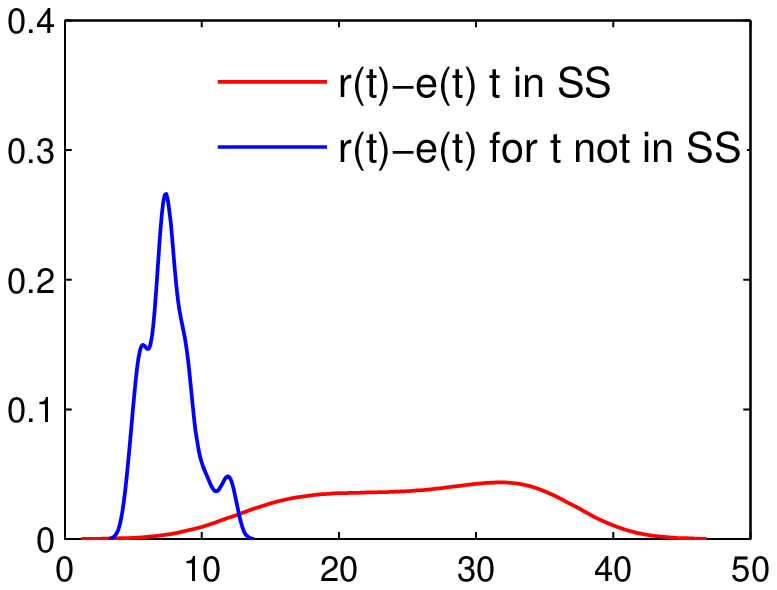}
        \caption{subject 2}
         \label{fig:portscan hist}
        \end{subfigure}~

       \begin{subfigure}[b]{0.22\textwidth}
        \includegraphics[width=\textwidth]{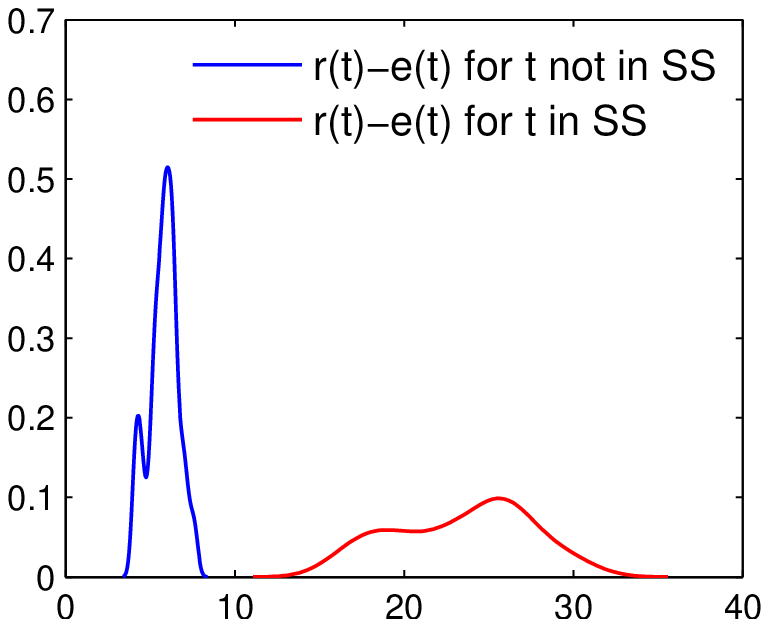}
        \caption{subject 3}
         \label{fig:portscan hist}
        \end{subfigure}~
       \begin{subfigure}[b]{0.22\textwidth}
        \includegraphics[width=\textwidth]{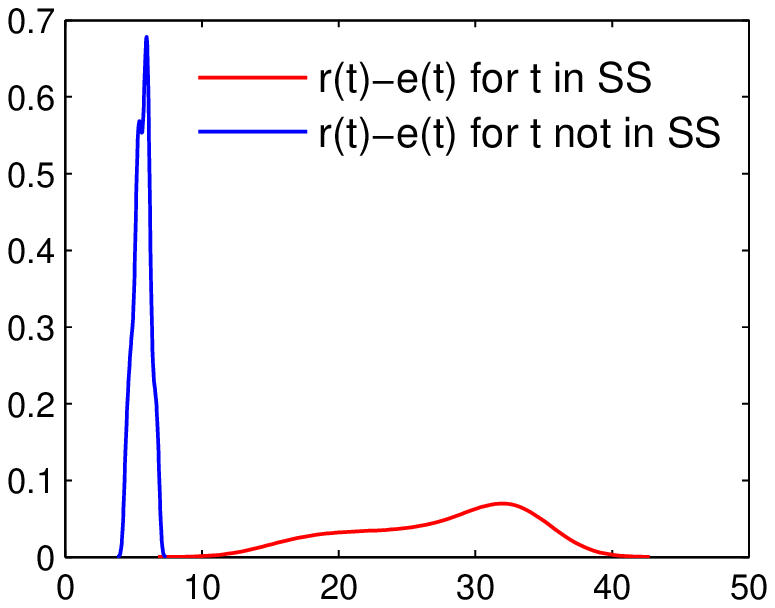}
        \caption{subject 4}
         \label{fig:portscan hist}
        \end{subfigure}
        \caption{Comparison of the distribution of the norm of $r(t)-e(t)$ for SS and
non-SS intervals. In all four subjects the designed residual suppresses spindles as designed as the
norm is higher for SS intervals.}
\label{fig:RE}
\end{figure}

\begin{figure}[ht!]
        \centering
         \includegraphics[width=0.4\textwidth]{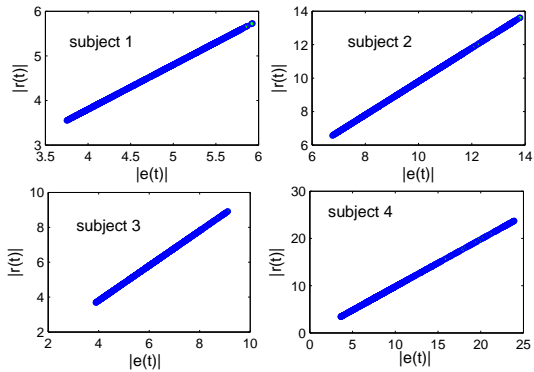}
        \caption{the $|r(t)|$ against $|e(t)|$ when $t$ is not in a spindle interval as  $r(t) = {\bf W}e(t)$}
\label{fig:RE2}
\end{figure}

In this section we evaluate whether the new residual $r(t)$ satisfies the design criterion.
Recall, $r(t)$ was designed to suppress the signal whenever a sleep spindle (SS) appears and
behave like the observed error ${\bf e(t)}$ in otherwise. Figure~\ref{fig:RE} shows the distribution
for $|r(t) - e(t)|_{2}$ for values of $t$ when $t$ is in (and not in) the predicted
SS interval $[a^{'}_{j},b^{'}_{j}]$ for some $j$. It is clear that the distribution when $t$ is in a predicted SS
interval is towards the right compared to when it is not in the interval. This is because in an SS interval, $r(t)$ will have a small absolute value (by
design). In a non-SS interval, $r(t)$ will be a linear function of $e(t)$, as  $r(t) = {\bf W}e(t)$. This
behavior is observed across subjects suggesting that in all cases that $r(t)$ is behaving as
designed. Furthermore in Figure~\ref{fig:RE2}, we plot the $|r(t)|$ against $|e(t)|$ when $t$ is
not in a spindle interval. Again we observe a straight line behavior, providing further confirmation
that $r(t)$ is behaving according to specifications.

\subsection{Delay in Detection of Anomalies}
\begin{figure} [ht!]
        \centering
        \begin{subfigure}[b]{0.24\textwidth}
                \includegraphics[width=\textwidth]{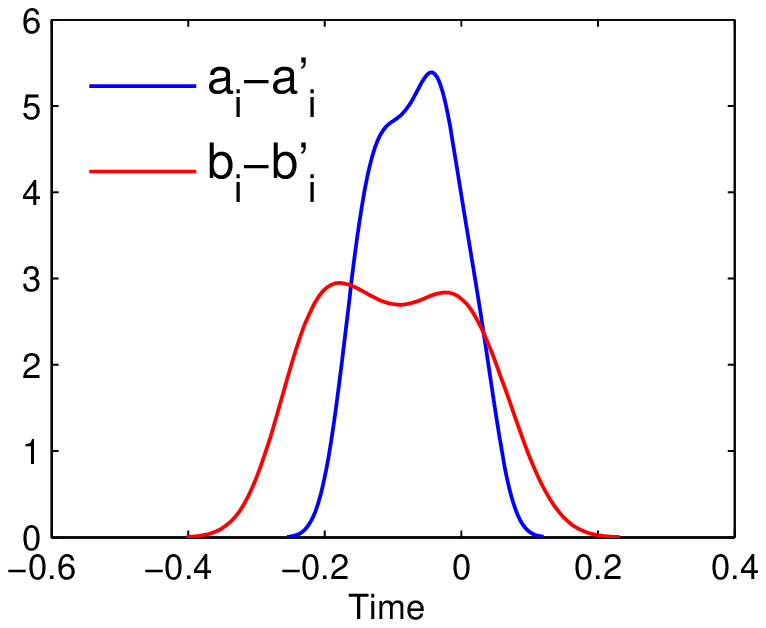}
                \caption{subject 1}
                \label{fig:DoS hist}
        \end{subfigure}~
       \begin{subfigure}[b]{0.24\textwidth}
        \includegraphics[width=\textwidth]{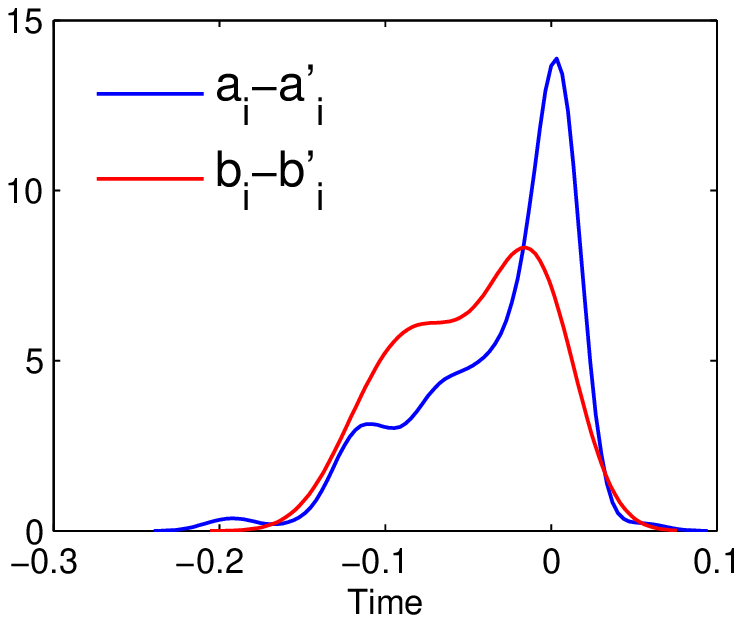}
        \caption{subject 4}
         \label{fig:portscan hist}
        \end{subfigure}~
         \caption{OSAD provides near real time detection. The delay between the actual appearance
of a spindle and the predicted appearance is a fraction of a second. Similarly the lag between
when the actual spindle disappears and it is reported to disappear is very small too. The x-axis
is in seconds.}
\label{fig:Delay_d}
\end{figure}

\begin{figure}
        \centering
        \includegraphics[width=0.45\textwidth]{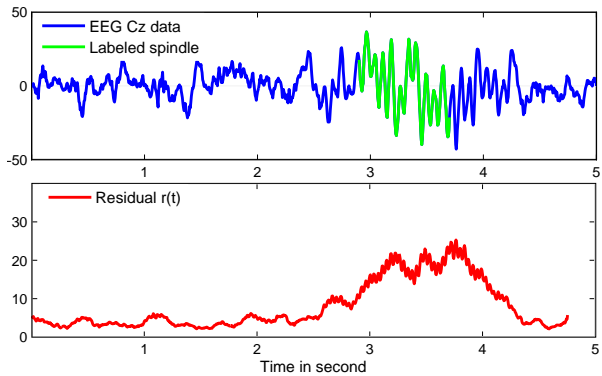}
        \vspace{10pt}
        \caption{Top: Cz data and a typical sleep spindle labeled. Bottom: Residual and detected sleep spindle.
In general
the predicted spindle interval is longer than the labeled interval. The predicted interval tends to include
the labeled interval, i.e., it begins earlier and finishes later. The  EEG shows that the labeled
intervals are actually quite conservative.}
\label{fig:ss}
\end{figure}

OSAD detects anomalies in near real time. We now discuss the lag between the appearance of a SS and
before it is reported by the LDS. Figure~\ref{fig:Delay_d} presents the delay distributions for subject 1 and subject 4
who experienced 164 and 132 labeled sleep spindles, respectively. In general, the predicted SS interval are
longer and contain the actual intervals.  This is confirmed in
Figure~\ref{fig:ss} which shows one specific example of
the location of the labeled sleep spindle and the predicted interval. In this case (which is typical),
the prediction of SS begins before and ends later than the labeled spindle.
Table~\ref{table:res7} shows the results of the mean delay between matched intervals.
Thus a mean of $(a_{i},a'_{i})$ equal to -0.0678 implies that on average, there was a delay of 1/200 second
before LDS reported an anomaly. On the other hand for subject 2 there the SS was, on average,
reported before it showed up in the labeled sequence. As noted in ~\cite{DRozario2013}, this is consistent
with the observation (and confirmed by double-blind scoring) that the labeling of SS is more conservative
i.e., SS are labeled for a shorter duration than what they should be.

\begin{table} [!t]
\caption{Delay statistics. The lag between appearance and prediction of SS is, on average, a fraction of a second.}
\centering
\vspace{-5pt}
\resizebox{.99\hsize}{!}{
\begin{tabularx} {0.7\textwidth}{ >{\bfseries \sffamily}r  p{4pt}  >{\sffamily}c p{1pt} >{\sffamily}c p{1pt} >{\sffamily}c  p{1pt} >{\sffamily}c }
\toprule
\addlinespace[12pt]
          &    &  Mean$(a_i-a'_i)$ & &  Mean$(b_i-b'_i)$&  &Std$(a_i-a'_i)$ & & Std$(b_i-b'_i)$ \\ [0.99ex]
                   \cmidrule{3-3}  \cmidrule{5-5} \cmidrule{7-7} \cmidrule{9-9} \\
\addlinespace[2pt]
subject 1 &    &  - 0.0678& & - 0.0961 &  &0.0589 & & 0.0995 \\ [0.99ex]
subject 2 &    &  0.0041 &  &0.0016  &  &0.0113 & & 0.0089 \\ [0.99ex]
subject 3 &    & -0.0426& & 0.0663 &  &0.0550 & &0.0478\\ [0.99ex]
subject 4 &    & -0.0340& & -0.0480 &  &0.0474 &  &0.0407 \\ [0.99ex]
\bottomrule
\end{tabularx}
}
\label{table:res7}
\end{table}

\section{Related work}
Automatic detection of sleep spindles is now an important topic in biomedical research. Different techniques
including FFTs, wavelet analysis and autoregressive time series modeling have been applied for sleep spindle
detection~\cite{Ray2010,Duman2009,Huupponen2007}. Attempts to integrate SVM to detect sleep spindles
have also been explored~\cite{Acir2005}. There seems to be a large variability between sleep EEG across
subjects. In our experiments we have  also observed this phenomenon. This combined with the large amount
of EEG noise has resulted in low level of agreement on the exact profile of sleep spindle~\cite{Nonclercq2013}.

The use of  Linear Dyamical Systems (LDS) to model time series is ubiquitous both in computer science \cite{Roweis1999}
and control theory \cite{Overschee1996,Ljung2001,Nelles2001,Cochrane2001,Kitagawa1987}. Expressing LDS in the language of graphical models and connections
with HMM have been extensively examined in machine learning. The use of LDS for anomaly detection
has also been investigated in network anomaly detection, among other areas~\cite{Soule2005}. The use of subspace identification methods for inferring the parameters
of LDS have been discussed by Overschee~\cite{Overschee1996}. Subspace methods estimate LDS parameters through a
spectral decomposition of a matrix of observations to yield an estimate of the underlying state space. Subspace
methods have low computational cost, are robust to perturbations  and are relatively easy to implement.
The recently introduced spectral learning methods are variations of the subspace method ~\cite{Boots2010,Hsu2008}

The use of eigenstructure assignment to alter the residual of an LDS has been investigated in the control
theory literature especially in the context of fault diagnosis~\cite{Andry1983}.Our approach closely
follows the work Patton et. al.~\cite{Patton1991} who have used eigenstructure assignment for altering the LDS model
using feedback. Other variations of LDS and fault diagnosis are discussed in ~\cite{Chen1996,Patton1997,Chen1999,Patton2000}.

\label{section:related}
\section{Conclusion}
In this paper we have introduced a new problem, the Online Selective Anomaly Detection (OSAD) to
capture a specific scenario in sleep research. Scientists working on sleep EEG data required
an alert system, which trigger alerts on selected anomalies. For example, sleep stage two
is characterized by two known anomalies: sleep spindle and K-complex. The requirement was
to design a system which detected both anomalies but only generated an alert when a non sleep
spindle anomaly appeared. We combined methods from data mining, machine learning and
control theory to design such a system. Experiments on real data set demonstrate that
our approach is accurate and produces the required results and is potentially applicable
to many other situations. We also note that data from sleep EEG provides a fertile ground
to apply existing data mining methodologies and potentially design new computational problems
and algorithms.

\section{Acknowledgments}
This work is partially supported by NICTA\footnote{http://nicta.com.au/}. NICTA is funded by the Australian Government as represented by the Department of Broadband, Communications and the Digital Economy and the Australian Research Council through the ICT Centre of Excellence program.

\vspace{70pt}
%
\bibliographystyle{abbrv}
\bibliography{sigproc}  
%
%
\vspace{160pt}

\appendix


\section{Proof of Theorem 1}
\label{app:proof}

\emph{Theorem 1. For a DRM, a sufficient condition for ${\bf G}_{\xi}(z)$ to be zero, is
}
\[
{\bf C}_f{\bf P}  = 0 \mbox{ and }
\{{\bf C}_f{\bf A}_f = 0 \mbox{ or } {\bf A}_f {\bf P} = 0\}
\]
\textbf{
Proof:} Let the set $\{\lambda_i=0,v_i\}$, for $i=1:n$, be the left eigenvectors and corresponding eigenvalues of ${\bf A}_f$, i.e.
\[
\begin{array}{rl}
                       v_i  {\bf A}_f &= \lambda_i v_i  \\
                       &= 0\\
\end{array}
\]
If one chooses $v_1$ as the rows of matrix $[{\bf WC}]$, then:
\[
\begin{array}{c}
                      \begin{bmatrix} v_1 &...&v_n \end{bmatrix}' {\bf A}_f = 0 \quad \Rightarrow \quad {\bf WC}{\bf A}_f = 0\\
\end{array}
\]

The matrix ${\bf A}_f={\bf A}-{\bf F}{\bf C}$, so it is sufficient to chose ${\bf F}$ so that the set $\{\lambda_i=0, v_i=[{\bf CW}]'\}$ to be assigned as left eigenpairs of $({\bf A}-{\bf F}{\bf C})$.\\

In the other side, suppose If the columns of ${\bf P}$ are the right eigenvectors of ${\bf A}_f$ corresponding to zero-values eigenvectors, then
\[
\begin{array}{c}
               {\bf A}_f v_i=0 \quad \Rightarrow \quad {\bf A}_f {\bf P} = 0\\
\end{array}
\]

So it is sufficient to chose ${\bf F}$ so that the set $\{\lambda_i=0, v_i={\bf P}\}$ to be assigned as right eigenpairs of $({\bf A}-{\bf F}{\bf C})$.\\


\end{document}